\documentclass[lettersize,journal]{IEEEtran}
\usepackage{amsmath,amsfonts}
\usepackage{array}
\usepackage[caption=false,font=normalsize,labelfont=sf,textfont=sf]{subfig}
\usepackage{stfloats}
\usepackage{url}
\usepackage{verbatim}
\usepackage{cite}
\usepackage{amsmath,amssymb,amsfonts}
\usepackage{algorithm,algorithmic}
\usepackage{graphicx}
\usepackage{textcomp}
\usepackage{xcolor}
\usepackage{caption}
\newcommand\norm[1]{\left\lVert#1\right\rVert}

\newcommand\abs[1]{\left\lvert #1\right\rvert}
\begin{document}

\title{Geometry Aware Meta-Learning Neural Network for Joint Phase and Precoder Optimization in RIS}

\author{\IEEEauthorblockN{%
        Dahlia Devapriya, %
        $\, $%
        Aparna V C, %
        $\, $%
        Sheetal Kalyani}\\
    \IEEEauthorblockA{\textit{Department of Electrical Engineering,} \\
        \textit{Indian Institute of Technology Madras,}
        \textit{
            Chennai - 600036, 
            India.
        }
        \\
        e-mail: \{ee22d003@smail, ee22d026@smail, skalyani@ee\}.iitm.ac.in}
        }



\maketitle

\begin{abstract}
In reconfigurable intelligent surface (RIS) aided systems, the joint optimization of the precoder matrix at the base station and the phase shifts of the RIS elements involves significant complexity. In this paper, we propose a complex-valued, geometry aware meta-learning neural network that maximizes the weighted sum rate in a multi-user multiple input single output system. By leveraging the complex circle geometry for phase shifts and spherical geometry for the precoder, the optimization occurs on Riemannian manifolds, leading to faster convergence. We use a complex-valued neural network for phase shifts and an Euler inspired update for the precoder network. Our approach outperforms existing neural network-based algorithms, offering higher weighted sum rates, lower power consumption, and significantly faster convergence. Specifically, it converges faster by nearly 100 epochs, with a 0.7 bps improvement in weighted sum rate and a 1.8 dB power gain when compared with existing work. Further it outperforms the state-of-the-art alternating optimization algorithm by 0.86 bps with a 2.6 dB power gain.
\end{abstract}


\section{Introduction}
\label{sec:intro}
Reconfigurable Intelligent Surfaces (RIS) 
will play an important role in emerging wireless systems and have been studied extensively in the last few years
\cite{pan2021reconfigurable}. A key area of research has been focusing on the fact that
in order to maximize the sum rate of the users, the precoder at the BS and the RIS phase shifts have to be optimized\cite{subhash2023optimal,nayak2024drl,zhu2024robust,chen2023integrated}. The phase shifts of the RIS elements have to be optimized with the unit modulus constraint, while the precoder has to satisfy the transmit power constraint. The joint optimization problem is non-convex and NP-hard. Early approaches used alternating optimization where one of the optimization variables is kept fixed while the other gets optimized\cite{huang2019reconfigurable, zhang2020sum,abeywickrama2020intelligent}. Specifically, the phase optimization is done using traditional iterative algorithms like Gradient Descent (GD) and the precoder optimization is inspired from the weighted sum MSE minimization (WMMSE) algorithm\cite{shi2011iteratively}. However, these approaches take longer to converge and involve costly matrix inversions, in addition to being sub-optimal. Hence recently neural networks were explored as a viable option. 

Deep reinforcement learning (DRL) based algorithms use deep deterministic policy gradient (DDPG) in order to perform the joint optimization\cite{huang2021multi,chen2023integrated,nayak2024drl}. In \cite{nayak2024drl}, a two-stage DRL based channel-oblivious algorithm is proposed. 
In \cite{xu2021robust}, a deep learning based algorithm is proposed for a system with discretized phase shifts and imperfect channel state information (CSI). Following this work, the authors of \cite{wang2024} proposed a meta-learning based joint optimization framework using a shallow neural network which improved upon the results of \cite{xu2021robust}. An additional constraint on the search space of the precoder is enforced in \cite{zhu2024robust} obtaining  state-of-the-art improved sum rates.

Although the neural network approaches have their merits, they do not exploit the inherent geometry of the optimization problem, namely, the underlying complex circle manifold corresponding to the phase shift constraint and the sphere manifold corresponding to the precoder power constraint. 
There are some non neural network based conventional methods which exploit complex circle manifold\cite{yu2019miso,li2020weighted,guo2020weighted}, however, they do not exploit the sphere manifold. A recent work (not neural network based) exploits both manifolds\cite{perovic2021achievable}. However, the corresponding precoder projection is highly computation intensive.

In this paper, we design a complex-valued geometry aware meta-learning neural network where, (a) the weights are updated according to the Riemannian ADAM manifold optimizer, (b) both the complex circle geometry and the spherical geometry of the RIS phase shifts and the precoder entries respectively are leveraged, (c) we propose a complex-valued neural network since the complex circle manifold incorporation requires a complex space and (d) a recent Euler equation based update proposed in machine learning literature\cite{zhang2019towards} is exploited for the precoder network design. Using the above complex-valued geometry aware neural network, we show significantly faster convergence and higher weighted sum rate against the state-of-the-art neural-network based algorithm\cite{zhu2024robust} as well a state-of-the-art traditional alternating optimization (AO) approach \cite{li2020weighted} for a multi-user multiple input single output (MU-MISO) system aided by an RIS. We name our approach as geometry aware meta learning neural network (GAMN).

\section{System model}
\label{sec:sys}
Consider a MU-MISO system consisting of a base station, an RIS, and $K$ single-antenna users. The base station consists of $M$ antennas and the RIS consists of $N$ reflective elements. Due to blockage, it is assumed that the direct BS-user link is not available for transmission. Hence the communication takes place through the BS-RIS-user link. We denote the channel between the BS and the RIS as $\mathbf{H}_{BR}\in\mathbb{C}^{N\times M}$ and the channel between the RIS and the $K$ users as $\mathbf{H}_{RU}\in\mathbb{C}^{K\times N}$. Each row of the matrix $\mathbf{H}_{RU}$ corresponds to the channel between the RIS and a user. We denote the RIS phase shift matrix as $\boldsymbol{\Theta}=\operatorname{diag}[e^{j\theta_1},e^{j\theta_2},\hdots,e^{j\theta_N}]\in \mathbb{C}^{N\times N}$ where $\theta_n$ corresponds to the phase shift by the $n^{th}$ RIS element. We denote the precoder matrix at the BS as $\mathbf{W}\in \mathbb{C}^{M\times K}$. Let $\mathbf{h}_{RU_k}\in\mathbb{C}^{N\times 1}$ denote the transpose of the $k^{th}$ row of the matrix $\mathbf{H}_{RU}$. Let $\mathbf{w}_{k}\in\mathbb{C}^{M\times 1}$ denote the $k^{th}$ column of the matrix $\mathbf{W}$. Let $s_k$ be the symbol transmitted to the $k^{th}$ user by the BS with $\mathbb{E}[\abs{s_k}^2]=1$ and let $P$ denote the total power at the BS. Hence the precoder matrix has to satisfy the constraint $\operatorname{tr}(\mathbf{W}\mathbf{W}^H)\leq P$. The downlink signal received by the $k^{th}$ user can be expressed as,
\begin{equation}\label{yk}
    y_k=\mathbf{h}_{RU_k}^H \boldsymbol{\Theta} \mathbf{H}_{BR} \mathbf{w}_k s_k +\sum_{i\ne k}^K \mathbf{h}_{RU_k}^H \boldsymbol{\Theta} \mathbf{H}_{BR} \mathbf{w}_i s_i +n_k,
\end{equation}
where $n_k \sim \mathcal{CN}(0,\sigma^2)$ which is the additive complex Gaussian noise of variance $\sigma^2$. The first term of \eqref{yk} denotes the signal corresponding to the target user while the second term denotes the interference term due to the other $K-1$ users. Hence the SINR at the $k^{th}$ user can be written as,
\begin{equation}\label{sinr}
    \gamma_k = \frac{\abs{\mathbf{h}_{RU_k}^H \boldsymbol{\Theta} \mathbf{H}_{BR} \mathbf{w}_k}^2}{\sigma^2 + \sum_{j\ne k}^K \abs{\mathbf{h}_{RU_k}^H \boldsymbol{\Theta} \mathbf{H}_{BR} \mathbf{w}_j}^2}.
\end{equation}
The weighted sum rate is widely used as a performance metric in several works\cite{shi2011iteratively,guo2020weighted,xu2021robust,xia2021meta,li2020weighted}. Using \eqref{sinr} it can be written as,
\begin{equation}
    R(\mathbf{W},\boldsymbol{\Theta},\mathbf{H}_{BR},\mathbf{H}_{RU})=\sum_{k=1}^K c_k \log_2(1+\gamma_k),
\end{equation}
where $c_k$ denotes the weight assigned to the rate corresonding to the $k^{th}$ user. These weights are initialized such that $\sum_{i=1}^K c_i=1$. We determine the optimal RIS phase shifts and precoder matrix which maximize the weighted sum rate. It can be written formally as,
\begin{align*} \textbf{P1:} \quad & \max_{\mathbf{W},\boldsymbol{\Theta}} \quad R(\mathbf{W},\boldsymbol{\Theta},\mathbf{H}_{BR},\mathbf{H}_{RU}) \\ \text{ s.t. } \quad & \operatorname{tr}(\mathbf{W}^H \mathbf{W}) \leq P, \\ & \boldsymbol{\Theta}=\operatorname{diag}[e^{j\theta_1},e^{j\theta_2},\hdots,e^{j\theta_N}], \\ &  0\leq\theta_k \leq 2\pi, k=1,\hdots,N. \end{align*} 
Due to the non-convexity and NP-hardness of \textbf{P1}, many classical approaches fail to give a satisfactory solution. We propose a neural network based algorithm in the succeeding section. 
\section{Proposed algorithm}
\label{sec:alg}
We propose a complex-valued geometry aware meta-learning neural network algorithm in order to jointly optimize the phase shifts of the RIS elements and the BS precoder matrix. It consists of an outer loop which we term the \textit{meta-learner} and two inner loops, one corresponding to the RIS phase shifts and the other corresponding to the precoder matrix which we term as the \textit{phase-learner} (PL) and the \textit{precoder-learner} (PRL) respectively.
\begin{figure}[tb]
\begin{minipage}[b]{1.0\linewidth}
  \centering
  \centerline{\includegraphics[width=8cm]{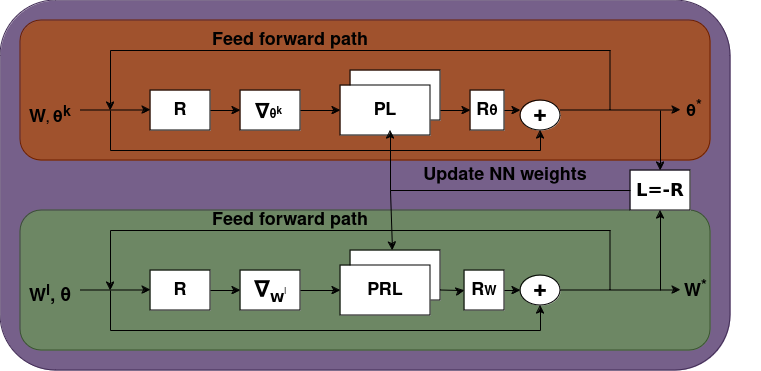}}
\end{minipage}
\caption{Block diagram view of the algorithm}
\label{fig:res}
\end{figure}
\subsection{Meta Learner}
The meta-learner minimizes the overall cost of the function i.e., the negative of the weighted sum rate and updates the weights of the two sub-networks namely the PL and the PRL accordingly through back propagation.
It receives the RIS phase shifts from the PL and the precoder matrix from the PRL and computes the loss function which is given by,
\begin{equation}
    \mathcal{L}=-R(\mathbf{W},\boldsymbol{\Theta},\mathbf{H}_{BR},\mathbf{H}_{RU}).
\end{equation}
We then back-propagate the gradient of the loss and update the weights of the PL and the PRL. We denote the weights of the PL and the PRL as $\mathbf{x}_P$ and $\mathbf{x}_{PR}$ respectively. 
First, we take a closer look at \textbf{P1} and the constraints of the problem. Let $\hat{\mathbf{w}}=[\mathbf{w}_1^T,\hdots,\mathbf{w}_K^T]^T$. The constraint $ \operatorname{tr}(\mathbf{W}^H \mathbf{W}) \leq P$ can be written as $\hat{\mathbf{w}}^H\hat{\mathbf{w}}\leq P$. 
We define the manifold, $$\mathcal{M}_{\mathbf{W}}=\{\mathbf{x}\in\mathbb{R}^{2MK}:\mathbf{x}^{T}\mathbf{x}=1\},$$ which is a manifold embedded in $\mathbb{R}^{2MK}$. The tangent space can be found from the kernel of the differential $Dh(\mathbf{x})[\mathbf{v}]= \lim_{t\rightarrow 0}\frac{(\mathbf{x}+t\mathbf{v})^T (\mathbf{x}+t\mathbf{v})}{t}=0$ as $\mathbf{x}^T\mathbf{v}=0$. Hence we can write the tangent space $\mathcal{T}_\mathbf{x} \mathcal{M}_W$ as, $$\mathcal{T}_\mathbf{x} \mathcal{M}_{\mathbf{W}}=\{\mathbf{v}\in \mathbb{R}^{2MK}:\mathbf{v}^{T}\mathbf{x}=<\mathbf{x},\mathbf{v}>=0\}.$$
Now, $\mathcal{M}_{\mathbf{W}}$ is a Riemannian sub-manifold of $\mathbb{R}^{2MK}$. Hence the Riemannian gradient of a function $f:\mathcal{M}_{\mathbf{W}}\rightarrow\mathbb{R}$ at $\mathbf{x}\in\mathbb{R}^{2MK}$ can be found by, $\operatorname{grad} f(\mathbf{x})=\nabla f(\mathbf{x}) -{\mathbf{x}}^T\nabla f(\mathbf{x})\mathbf{x}$.
The retraction $\mathcal{R}_{\mathbf{W}}:\mathcal{T}_\mathbf{x}\mathcal{M}_{\mathbf{W}}\rightarrow\mathcal{M}_{\mathbf{W}}$ from the tangent space to the sphere manifold can be done by the operation 
\begin{equation}\label{projw}
    \mathcal{R}_{\mathbf{W}}(v)=\frac{x+v}{\norm{x+v}}.
\end{equation}
Consider the complex circle manifold given by,$$\mathcal{M}_{\boldsymbol{\theta}}=\{x | x\in\mathbb{C},\abs{x}=1\}.$$ It lies in a space defined by the equation $h(x)=\abs{x}-1=0$. In order to find its tangent space we set the differential to zero as, $Dh(x)[v]=\lim_{t\rightarrow 0} \frac{(x+tv)^*(x+tv)-x^*x}{t}=0$ obtaining $v^*x=0$.
Hence the tangent space is given by, $$\mathcal{T}_{x}\mathcal{M}_{\boldsymbol{\theta}}=\{v | v^*x=0\}.$$ The Riemannian gradient of a function $f:\mathcal{M}_{\boldsymbol{\theta}}\rightarrow\mathbb{R}$ can be found by, $\operatorname{grad}f(x)=\frac{\partial f(x)}{\partial x} -{x}^*\frac{\partial f(x)}{\partial x}x$.
The retraction to the manifold $\mathcal{M}_{\boldsymbol{\theta}}$ is given by,
\begin{equation}\label{projt}
    \mathcal{R}_{\boldsymbol{\theta}}(v)=\frac{x+v}{\abs{x+v}}.
\end{equation}
Since the values of the phase shift matrix are constrained to be unit modulus and the corresponding values of the angles to lie in $[0,2\pi]$, we perform Riemannian optimization over $N$ complex circle manifolds corresponding to the $N$ RIS elements. Using the appropriate gradients, we perform Riemannian ADAM (RADAM) optimization according to the procedure in \cite{becigneulriemannian}. Hence now the optimization problem \textbf{P1} becomes,
\begin{align*} \textbf{P2:} \quad & \max_{\exp\{\boldsymbol{\theta}\}\in\mathcal{M}_{\boldsymbol{\theta}},\hat{\mathbf{w}}\in\mathcal{M}_W} \quad R(\mathbf{W},\boldsymbol{\Theta},\mathbf{H}_{BR},\mathbf{H}_{RU}) .\end{align*}
Let $t$ denote the index of the outer loop iteration. At each outer loop iteration, the updated weights are given by,
\begin{equation}\label{eq:phase}
    \mathbf{x}_P^{t+1}=\mathbf{x}_P^t -\alpha_P \text{RADAM}\left(\mathbf{x}_P^t, \nabla_{\mathbf{x}_P^t} \mathcal{L}^t\right)
\end{equation}
\begin{equation}\label{eq:precoder}
    \mathbf{x}_{PR}^{t+1}=\mathbf{x}_{PR}^t -\alpha_{PR} \text{RADAM}\left(\mathbf{x}_{PR}^t, \nabla_{\mathbf{x}_{PR}^t} \mathcal{L}^t\right),
\end{equation}
where $\alpha_P$ and $\alpha_{PR}$ denote the learning rates of the PL and the PRL networks respectively. This step is repeated for $n_M$ times in the outer loop. Note that \eqref{eq:phase} has complex values as the PL is a complex-valued neural network.
\subsection{Phase Learner}
We construct a complex-valued neural network consisting of a single hidden layer of $200$ neurons. The sizes of the input and the output layers are set equal to the number of RIS elements $N$. In \cite{zhu2024robust}, the authors show that the search for the local minima in the gradient space is more efficient compared to the search in the entire feasible region. Hence the input to the PL network consists of the gradient of the weighted sum rate with respect to the phase shifts $\boldsymbol{\theta}=[\theta_1,\hdots,\theta_N]^T\in\mathbb{C}^{N\times 1}$ i.e, $\nabla_{\boldsymbol{\theta}} R(\mathbf{W},\boldsymbol{\Theta},\mathbf{H}_{BR},\mathbf{H}_{RU})$. 

We then perform the feed-forward step,
\begin{equation}
    \boldsymbol{\theta}^{k+1} =\boldsymbol{\theta}^{k}+\text{PL}\left( \nabla_{\boldsymbol{\theta}} R(\mathbf{W},\boldsymbol{\Theta},\mathbf{H}_{BR},\mathbf{H}_{RU}), \mathbf{x}_P \right). 
\end{equation}
The feed-forward operation is repeated in the inner loop for $n_P$ times. Finally, we do a projection to the complex circle manifolds according to \eqref{projt}. 
\subsection{Precoder Learner}
The precoder network takes as input the gradient of the weighted sum rate with respect to the precoder weights which is flattened as a vector, separating its real and imaginary components. It consists of a hidden layer of $200$ neurons and outputs the precoder matrix.
The feed-forward update equation for the PL network is given by,
\begin{equation}
    \mathbf{W}^{l+1}=\mathbf{W}^{l} + h*PRL\left( \nabla_{\mathbf{W}} R(\mathbf{W},\boldsymbol{\Theta},\mathbf{H}_{BR},\mathbf{H}_{RU}), \mathbf{x}_{PR} \right),
\end{equation}
where $h$ denotes the Euler factor inspired by the explicit Euler method to solve partial differential equations (PDSs). Specifically, the solution to the PDE $\frac{\partial \mathbf{x}}{\partial t}=f(\mathbf{x},t)$ is given by,
\begin{equation}
    \mathbf{x}_{n+1}=\mathbf{x}_n+h*f(\mathbf{x}_n,t_n),
\end{equation}
where we write $\mathbf{x}_n(t_n)$ as $\mathbf{x}_n$. The Euler method was mapped to the success of robust Resnets in \cite{zhang2019towards}. Hence we include the Euler factor in our precoder network. We perform the feed-forward operation for $n_{PR}$ times after which we do a projection to the sphere manifold according to \eqref{projw}. Note that in contrast to \cite{perovic2021achievable} where the retraction is done by the water-filling algorithm after every update, we ensure our iterates stay on the spherical manifold by exploiting Riemannian geometry. This helps our algorithm to converge significantly faster with improved sum rates.

We give a block diagram view of the Meta-learner (the outer loop minimizing the overall cost), the phase network and the precoder network in Fig. \ref{fig:res}. We give our overall algorithm in Algorithm \ref{algo:gmn}. While the precoder is updated in every iteration, the phase shifts are updated once in every $n_{I}$ iterations in order to maintain the stability similar to \cite{zhu2024robust}.
  \begin{algorithm}
        \caption{GAMN Algorithm}
        \begin{algorithmic}[1]\label{algo:gmn}
            \STATE \textit{\textbf{Initialization}}: Randomly initialize $\mathbf{x}_P^0$, $\mathbf{x}_{PR}^0$, $\mathbf{W}^0$, $\boldsymbol{\theta}^0$ and set $\mathbf{W}^* = \mathbf{W}^0$ and $\boldsymbol{\theta}^*=\boldsymbol{\theta}^0$.
            \FOR{$t=1\text{ to }n_M$}
            \FOR{$k=1\text{ to }n_P$}
         \STATE  $R_{\boldsymbol{\theta}}^{k-1}=R(\mathbf{W}^*,\boldsymbol{\Theta}^{k-1},\mathbf{H}_{BR},\mathbf{H}_{RU})$
         \STATE $\Delta\boldsymbol{\theta}^{k-1}=\text{PL}\left( \nabla_{\boldsymbol{\theta}} R_{\boldsymbol{\theta}}^{k-1}, \mathbf{x}_P^{t-1} \right)$
         \STATE $\boldsymbol{\theta}^k = \boldsymbol{\theta}^{k-1}+\Delta\boldsymbol{\theta}^{k-1}$
         \ENDFOR
         \STATE $\boldsymbol{\theta}^{n_P}=\mathcal{R}_{\mathcal{\theta}}(\boldsymbol{\theta}^{n_P})$
         \STATE $\boldsymbol{\theta}^*=\boldsymbol{\theta}^{n_P}$
         \FOR{$l=1\text{ to }n_{PR}$}
         \STATE $R_{\mathbf{W}}^{l-1}=R(\mathbf{W}^{l-1},\boldsymbol{\Theta}^{*},\mathbf{H}_{BR},\mathbf{H}_{RU})$
         \STATE $\Delta \mathbf{W}^{l-1}=\text{PRL}\left( \nabla_{\mathbf{W}} R_{\mathbf{W}}^{l-1}, \mathbf{x}_{PR}^{t-1} \right)$
         \STATE $\mathbf{W}^l=\mathbf{W}^{l-1}+h\Delta \mathbf{W}^{l-1}$
         \ENDFOR
         \STATE $\mathbf{W}^{n_{PR}}=\mathcal{R}_{\mathbf{W}}(\mathbf{W}^{n_{PR}})$
         \STATE $\mathbf{W}^*=\mathbf{W}^{n_{PR}}$
         \STATE $\mathcal{L}^{t-1}=-R(\mathbf{W}^*,\boldsymbol{\Theta}^{*},\mathbf{H}_{BR},\mathbf{H}_{RU})$
         \STATE $\mathbf{x}_{PR}^t=\mathbf{x}_{PR}^{t-1}-\alpha_{PR}\text{RADAM}\left(\mathbf{x}_{PR}^{t-1}, \nabla_{\mathbf{x}_{PR}^{t-1}} \mathcal{L}^{t-1}\right)$
         \IF{$t\%n_I =0$}
         \STATE $\mathbf{x}_{P}^t=\mathbf{x}_{P}^{t-1}-\alpha_{P}\text{RADAM}\left(\mathbf{x}_{P}^{t-1}, \nabla_{\mathbf{x}_{P}^{t-1}} \mathcal{L}^{t-1}\right)$
         \ENDIF
           \ENDFOR
        \end{algorithmic}
    \end{algorithm}

\section{Results}
For our simulations, we set the same channel parameters and setup as in \cite{zhu2024robust} and \cite{wang2024}. The channel between the BS and the RIS as well as between RIS and the users is assumed to follow Rician fading.
Specifically,
\begin{equation}   \mathbf{h}_{RU_k}=L_{RU_k}^{LoS}\sqrt{\frac{\kappa_{RU}}{1+\kappa_{RU}}} \mathbf{h}_{RU_k}^{LoS} + L_{RU_k}^{NLoS}\sqrt{\frac{1}{1+\kappa_{RU}}}\mathbf{h}_{RU_k}^{NLoS}
\end{equation}
\begin{equation}
    \mathbf{H}_{BR}=L_{BR}^{LoS} \sqrt{\frac{\kappa_{BR}}{1+\kappa_{BR}}} \mathbf{H}_{BR}^{LoS} +L_{BR}^{NLoS}\sqrt{\frac{1}{1+\kappa_{BR}}} \mathbf{H}_{BR}^{NLoS},
\end{equation}
where $L_{RU_k}^{LoS}$, $L_{BR}^{LoS}$ and $L_{RU_k}^{NLoS}$, $L_{BR}^{NLoS}$ denote the path loss for the line of sight (LoS) and non line of sight (NLoS) components respectively. $\kappa_{RU}$ and $\kappa_{BR}$ are the Rician coefficients for the RIS-user and BS-RIS links respectively and are set equal to $10$. The BS is located at $(0,10)$ m, the users are randomly located in a circle of radius $5$ m centered at $(100,15)$ m, and the RIS at $(100,0)$ m. The path loss is set according to the 3GPP standard for LoS and NLoS components \cite{3gpp}. The BS antenna spacing is set at $0.5\lambda$ with $28GHz$ frequency. We set the parameters $n_M$, $n_P$, $n_{PR}$, $\alpha_{P}$ and $\alpha_{PR}$ as $500$, $1$, $1$, $10^{-2}$ and $3.5\times 10^{-2}$ respectively. We also run the algorithm for $100$ channel realizations and average our results. The default values of power, N, and M, unless otherwise specified, are 10 dB, 100, and 64, respectively.
\begin{figure}
    \centering
    \includegraphics[scale=0.6]{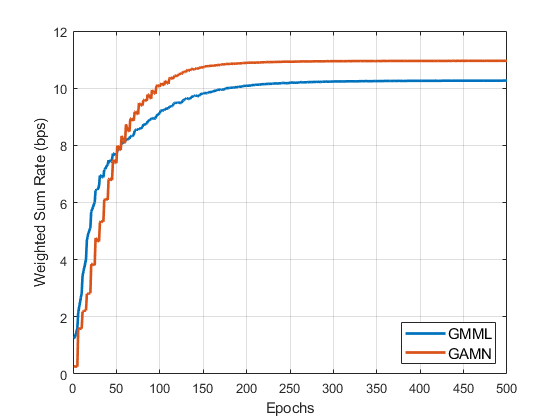}
    \caption{Weighted Sum Rate across the iterations}
    \label{fig:conv2}
\end{figure}

In Fig. \ref{fig:conv2}, we show the convergence behavior of our algorithm and compare it with the GMML algorithm, with the total power set to 10 dBm and the Euler factor appropriately tuned. We observe that GAMN converges nearly 100 epochs faster than GMML, owing to the manifold-based optimization and the incorporation of the Euler factor in our algorithm. Optimization on manifolds reduces the search space and accelerates convergence toward the global minimum, which explains the improved convergence rate. The traditional optimization technique for this problem employs an AO approach, where the beamforming matrix is optimized using the WMMSE algorithm and the phase shifts are optimized via the Riemannian conjugate gradient (RCG) algorithm \cite{li2020weighted}. This results in a multi-stage iterative process in which each subproblem requires its own iterative update procedure. In particular, the WMMSE algorithm involves matrix inversions in every update step, further increasing the computational complexity. The per-iteration computational cost and therefore the overall time complexity of AO is not comparable to that of GAMN or GAMML. Consequently, we do not include a convergence plot for AO in this setting. Note that AO converges within 100 iterations to a value of 10.0844, whereas GMML and GAMN surpass this value at epochs 200 and 100, respectively, with final converged values of 10.2594 and 10.9526.

\begin{figure}
    \centering
    \includegraphics[scale=0.6]{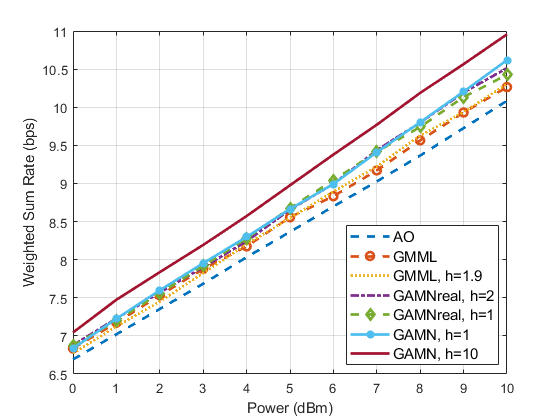}
    \caption{Weighted Sum Rate v/s Power}
    \label{fig:P}
\end{figure}

In Fig. \ref{fig:P}, we plot the weighted sum rate versus the total transmit power for $N=100$ and $M=64$. We compare the proposed GAMN algorithm with the state-of-the-art Gradient-based Manifold Meta-Learning (GMML) algorithm proposed in \cite{zhu2024robust}, as well as the traditional alternating optimization method \cite{li2020weighted}. To evaluate the impact of the complex-valued PL network, we also include a version of GAMN in which the PL network uses real-valued weights, denoted as GAMNreal. Furthermore, we incorporate the Euler factor into all the aforementioned algorithms and tune their parameters for optimal performance. From the plot, we observe that GAMN with the Euler factor set to $10$ achieves the best performance across all power levels. In addition, although the value of $h$ significantly improves the performance of GAMN, it does not yield substantial improvements for GMML or GAMNreal.

\begin{figure}
    \centering
    \includegraphics[scale=0.6]{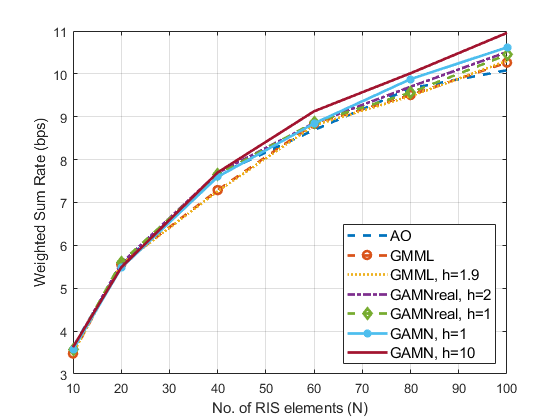}
    \caption{Weighted Sum Rate v/s number of RIS elements}
    \label{fig:N}
\end{figure}
\begin{figure}
    \centering
    \includegraphics[scale=0.6]{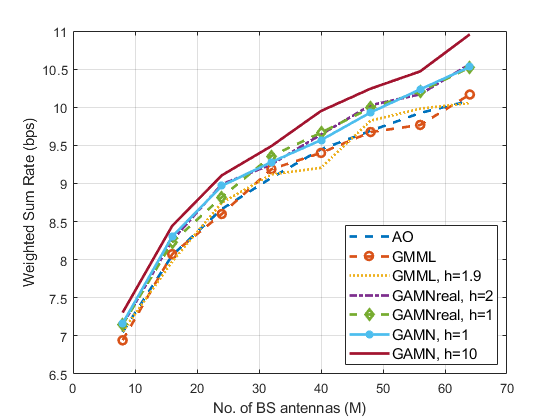}
    \caption{Weighted Sum Rate v/s number of BS antennas}
    \label{fig:M}
\end{figure}

In Fig. \ref{fig:N} and Fig. \ref{fig:M}, we plot the weighted sum rate versus the number of RIS elements ($N$) and the number of BS antennas ($M$). We clearly observe that the proposed GAMN method with an Euler factor of 10 outperforms both the state-of-the-art GMML method and the traditional optimization approach. Moreover, the proposed complex-valued PL network, when combined with an appropriately tuned Euler factor, achieves the best performance for different values of the number of RIS elements $N$ and the number of BS antennas $M$.


We observe $1.8$ dB of power gain from Fig. \ref{fig:P} comparing GAMN with $h=10$ and GMML. GAMN outperforms GMML by a weighted sum rate difference of $0.7$ bps with $10$ dBm power. Furthermore, GAMN outperforms the traditional AO baseline by $0.86$ bps at $10$ dBm power and a power gain of $2.6$ dB.


\textbf{Complexity analysis: }
The computational cost of the SINR evaluation is dominated by the term \( h_k^{H} \Theta G w_j \), whose computation requires \( O(K^{2} M N) \) multiplications. When this cost is combined with the iterative structure of the algorithm, where \( N_e \) denotes the number of epochs, \( N_o \) the number of outer iterations per epoch, and \( N_i \) the number of inner iterations within each outer iteration, the total computational complexity becomes \( O(N_e N_o N_i K^{2} M N) \), which is the same as for \cite{zhu2024robust}. Our choice of optimizer introduces no further computational overhead beyond this dominant term. Importantly, no cubic-order terms appear in the complexity expression, since the method does not rely on matrix inversions as in alternating optimization \cite{li2020weighted}. 


\section{Conclusion}
We propose a complex-valued geometry aware meta learning neural network which was able to outperform the state-of-the-art across multiple metrics. By identifying the inherent geometry in the problem and then using complex-valued neural networks to exploit it, we are able to get an improvement in sum rate, power, and computation cost. To the best of our knowledge complex-valued neural networks have not been used in communication applications despite baseband signals being complex. We hope this work will inspire complex-valued neural network designs that are also geometry aware in problems pertaining to wireless communication.

\bibliographystyle{IEEEtran}
\bibliography{paper}

@inproceedings{wang2024,
author = {Wang, Xinquan and Zhu, Fenghao and Zhou, Qianyun and Yu, Qihao and Chongwen, Huang and Alhammadi, Ahmed and Zhang, Zhaoyang and Yuen, Chau and Debbah, Mérouane},
booktitle={2024 IEEE International Conference on Communications},
year = {2024},
pages={3464--3469},
title = {Energy-efficient Beamforming for {RIS}-aided Communications: Gradient Based Meta Learning}
}

@article{pan2021reconfigurable,
  title={Reconfigurable intelligent surfaces for 6G systems: Principles, applications, and research directions},
  author={Pan, Cunhua and Ren, Hong and Wang, Kezhi and Kolb, Jonas Florentin and Elkashlan, Maged and Chen, Ming and Di Renzo, Marco and Hao, Yang and Wang, Jiangzhou and Swindlehurst, A Lee and others},
  journal={IEEE Communications Magazine},
  volume={59},
  number={6},
  pages={14--20},
  year={2021},
  publisher={IEEE}
}

@article{shi2011iteratively,
  title={An iteratively weighted MMSE approach to distributed sum-utility maximization for a {MIMO} interfering broadcast channel},
  author={Shi, Qingjiang and Razaviyayn, Meisam and Luo, Zhi-Quan and He, Chen},
  journal={IEEE Transactions on Signal Processing},
  volume={59},
  number={9},
  pages={4331--4340},
  year={2011},
  publisher={IEEE}
}

@article{guo2020weighted,
  title={Weighted sum-rate maximization for reconfigurable intelligent surface aided wireless networks},
  author={Guo, Huayan and Liang, Ying-Chang and Chen, Jie and Larsson, Erik G},
  journal={IEEE Transactions on Wireless Communications},
  volume={19},
  number={5},
  pages={3064--3076},
  year={2020},
  publisher={IEEE}
}

@article{abeywickrama2020intelligent,
  title={Intelligent reflecting surface: Practical phase shift model and beamforming optimization},
  author={Abeywickrama, Samith and Zhang, Rui and Wu, Qingqing and Yuen, Chau},
  journal={IEEE Transactions on Communications},
  volume={68},
  number={9},
  pages={5849--5863},
  year={2020},
  publisher={IEEE}
}

@article{zhang2020sum,
  title={Sum rate optimization for two way communications with intelligent reflecting surface},
  author={Zhang, Yu and Zhong, Caijun and Zhang, Zhaoyang and Lu, Weidang},
  journal={IEEE Communications Letters},
  volume={24},
  number={5},
  pages={1090--1094},
  year={2020},
  publisher={IEEE}
}

@article{huang2019reconfigurable,
  title={Reconfigurable intelligent surfaces for energy efficiency in wireless communication},
  author={Huang, Chongwen and Zappone, Alessio and Alexandropoulos, George C and Debbah, M{\'e}rouane and Yuen, Chau},
  journal={IEEE Transactions on Wireless Communications},
  volume={18},
  number={8},
  pages={4157--4170},
  year={2019},
  publisher={IEEE}
}

@article{huang2021multi,
  title={Multi-hop {RIS}-empowered terahertz communications: A {DRL}-based hybrid beamforming design},
  author={Huang, Chongwen and Yang, Zhaohui and Alexandropoulos, George C and Xiong, Kai and Wei, Li and Yuen, Chau and Zhang, Zhaoyang and Debbah, M{\'e}rouane},
  journal={IEEE Journal on Selected Areas in Communications},
  volume={39},
  number={6},
  pages={1663--1677},
  year={2021},
  publisher={IEEE}
}

@inproceedings{chen2023integrated,
  title={Integrated Beamforming and Resource Allocation in {RIS}-Assisted mmWave Networks based on Deep Reinforcement Learning},
  author={Chen, Di and Gao, Hui and Chen, Na and Cao, Ruohan},
  booktitle={2023 21st IEEE Interregional NEWCAS Conference},
  pages={1--5},
  year={2023},
  organization={IEEE}
}

@article{nayak2024drl,
  title={A {DRL} Approach for {RIS}-Assisted Full-Duplex {UL} and {DL} Transmission: Beamforming, Phase Shift and Power Optimization},
  author={Nayak, Nancy and Kalyani, Sheetal and Suraweera, Himal A},
  journal={IEEE Transactions on Wireless Communications, Early access},
  year={2024},
  publisher={IEEE}
}

@article{xu2021robust,
  title={A robust deep learning-based beamforming design for {RIS}-assisted multiuser {MISO} communications with practical constraints},
  author={Xu, Wangyang and Gan, Lu and Huang, Chongwen},
  journal={IEEE Transactions on Cognitive Communications and Networking},
  volume={8},
  number={2},
  pages={694--706},
  year={2021},
  publisher={IEEE}
}

@inproceedings{xia2021meta,
  title={Meta-learning based beamforming design for {MISO} downlink},
  author={Xia, Jingyuan and Gunduz, Deniz},
  booktitle={2021 IEEE International Symposium on Information Theory},
  pages={2954--2959},
  year={2021},
  organization={IEEE}
}

@article{perovic2021achievable,
  title={Achievable rate optimization for {MIMO} systems with reconfigurable intelligent surfaces},
  author={Perovi{\'c}, Nemanja Stefan and Tran, Le-Nam and Di Renzo, Marco and Flanagan, Mark F},
  journal={IEEE Transactions on Wireless Communications},
  volume={20},
  number={6},
  pages={3865--3882},
  year={2021},
  publisher={IEEE}
}

@article{li2020weighted,
  title={Weighted sum-rate maximization for multi-{IRS} aided cooperative transmission},
  author={Li, Zhengfeng and Hua, Meng and Wang, Qingxia and Song, Qingheng},
  journal={IEEE Wireless Communications Letters},
  volume={9},
  number={10},
  pages={1620--1624},
  year={2020},
  publisher={IEEE}
}

@inproceedings{yu2019miso,
  title={{MISO} wireless communication systems via intelligent reflecting surfaces},
  author={Yu, Xianghao and Xu, Dongfang and Schober, Robert},
  booktitle={2019 IEEE/CIC International Conference on Communications in China},
  pages={735--740},
  year={2019},
  organization={IEEE}
}

@article{zhu2024robust,
author = {Zhu, Fenghao and Wang, Xinquan and Chongwen, Huang and Yang, Zhaohui and Chen, Xiaoming and Alhammadi, Ahmed and Zhang, Zhaoyang and Yuen, Chau and Debbah, Mérouane},
year = {2024},
title = {Robust Beamforming for {RIS}-aided Communications: Gradient-based Manifold Meta Learning},
journal = {IEEE Transactions on Wireless Communications, Early access},
doi = {10.1109/TWC.2024.3435023}
}

@article{3gpp,
  title = {Further Advancements for E-UTRA Physical Layer Aspects (Release 9)},
  author = {3GPP},
  journal = {TS 36.814},
  volume = {V9.2.0},
  issue = {},
  pages = {},
  numpages = {},
  year = {2010},
  month = {March},
  publisher = {}
}

@inproceedings{becigneulriemannian,
  title={Riemannian Adaptive Optimization Methods},
  author={Becigneul, Gary and Ganea, Octavian Eugen},
  booktitle={International Conference on Learning Representations},
year={2019}
}

@inproceedings{zhang2019towards,
  title={Towards robust ResNet: a small step but a giant leap},
  author={Zhang, Jingfeng and Han, Bo and Wynter, Laura and Low, Bryan Kian Hsiang and Kankanhalli, Mohan},
  booktitle={Proceedings of the 28th International Joint Conference on Artificial Intelligence},
  pages={4285--4291},
  year={2019}
}

@article{subhash2023optimal,

  author={Subhash, Athira and Kammoun, Abla and Elzanaty, Ahmed and Kalyani, Sheetal and Al-Badarneh, Yazan H. and Alouini, Mohamed-Slim},

  journal={IEEE Journal on Selected Areas in Communications}, 

  title={Optimal Phase Shift Design for Fair Allocation in {RIS}-Aided Uplink Network Using Statistical {CSI}}, 

  year={2023},

  volume={41},

  number={8},

  pages={2461-2475}}
\end{document}